\documentclass[10pt,twocolumn,letterpaper]{article}

\usepackage{iccv}
\usepackage{times}
\usepackage{epsfig}
\usepackage{graphicx}
\usepackage{amsmath}
\usepackage{amssymb}
\usepackage{tabularx}
\usepackage{booktabs}
\usepackage{MnSymbol}
\usepackage{multirow}
\usepackage[skip=0.3\baselineskip]{caption}
\usepackage{subcaption}
\usepackage[export]{adjustbox}
\usepackage{bbm}
\usepackage{bm}
\usepackage{makecell}

\usepackage{algorithm}
\usepackage[noend]{algpseudocode}

\usepackage[pagebackref=true,breaklinks=true,letterpaper=true,colorlinks,bookmarks=false]{hyperref}

\iccvfinalcopy 

\def\iccvPaperID{3861} 

\ificcvfinal\fi

\begin{document}

\title{N-ImageNet: Towards Robust, Fine-Grained \protect \\ Object Recognition with Event Cameras}

\author{Junho Kim, Jaehyeok Bae, Gangin Park, Dongsu Zhang, and Young Min Kim\\
Dept. of Electrical and Computer Engineering, Seoul National University, Korea\\
{\tt\small \{82magnolia, wogur110, ssonpull519, 96lives, youngmin.kim\}@snu.ac.kr}
}

\maketitle
\ificcvfinal\thispagestyle{empty}\fi

\begin{abstract}
We introduce N-ImageNet, a large-scale dataset targeted for robust, fine-grained object recognition with event cameras.
The dataset is collected using programmable hardware in which an event camera consistently moves around a monitor displaying images from ImageNet.
N-ImageNet serves as a challenging benchmark for event-based object recognition, due to its large number of classes and samples.
We empirically show that pretraining on N-ImageNet improves the performance of event-based classifiers and helps them learn with few labeled data.
In addition, we present several variants of N-ImageNet to test the robustness of event-based classifiers under diverse camera trajectories and severe lighting conditions, and propose a novel event representation to alleviate the performance degradation.
To the best of our knowledge, we are the first to quantitatively investigate the consequences caused by various environmental conditions on event-based object recognition algorithms.
N-ImageNet and its variants are expected to guide practical implementations for deploying event-based object recognition algorithms in the real world.
Code is available at \url{https://github.com/82magnolia/n_imagenet/}.

\end{abstract}

\section{Introduction}
Event cameras are neuromorphic vision sensors that encode visual information as a sequence of events, and have a myriad of benefits such as high dynamic range, low energy consumption, and microsecond-scale temporal resolution.
However, algorithms for processing event data are still at their nascency.
This is primarily due to the lack of a large, fine-grained dataset for training and evaluating different event-based vision algorithms.
While the number of event camera datasets surged in the past few years, many fine-grained datasets lack size~\cite{n_caltech}, whereas large-scale real-world datasets lack label diversity~\cite{hats}.
Large amounts of publicly available data are one of the key factors in the recent success of computer vision.
For example, ImageNet~\cite{imagenet} triggered the development of accurate, high performance object recognition algorithms~\cite{resnet, alexnet} whereas MS-COCO~\cite{coco} led to the advent of eloquent image captioning systems~\cite{show_attend_and_tell}.

\begin{figure}
\centering
\subfloat{%
  \includegraphics[width=\columnwidth]{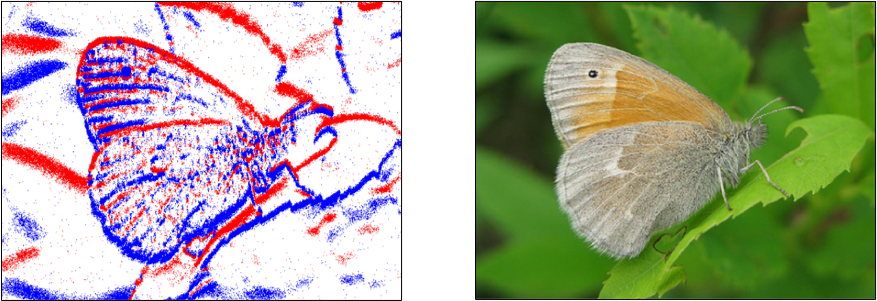}%
}
\\[+2ex]
\subfloat{%
  \includegraphics[width=\columnwidth]{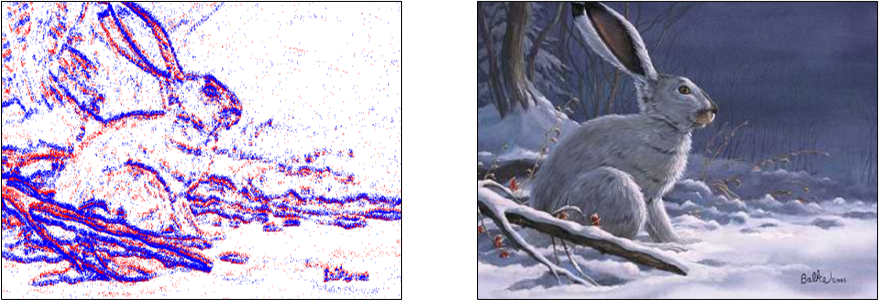}%
}
\caption{Sample events from N-ImageNet displayed along with their RGB counterparts from ImageNet~\cite{imagenet}. Positive, negative events are shown in blue and red, respectively.}
\label{device}
\end{figure}

\begin{figure}
\centering
\includegraphics[width=\columnwidth]{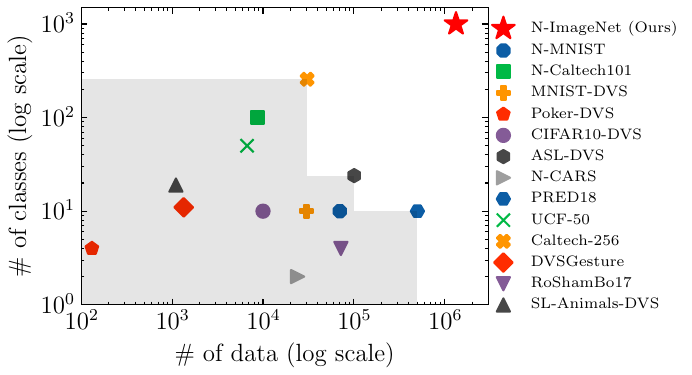}
\caption{Comparison of N-ImageNet (the red star in the upper-right corner) against existing datasets (distributed in the area shaded in gray) in terms of dataset size and class count. Note that the axes are displayed in log scale.}
\label{fig:dataset_stats}
\end{figure}

We provide N-ImageNet, an event camera dataset targeted for object recognition that surpasses all existing datasets in both size and label granularity as summarized in Table~\ref{b_as_b} and Figure~\ref{fig:dataset_stats}.
Since it is infeasible to manually obtain real-world instances of thousands of object categories,
we opt to generate events by moving the sensor in front of an LCD monitor which displays images from ImageNet~\cite{imagenet} as in \cite{n_caltech, cifar_dvs} using programmable hardware.

N-ImageNet is projected to function as a challenging benchmark for event-based object recognition algorithms.
As shown in Table~\ref{b_as_b}, evaluations of various classifiers on N-ImageNet demonstrate a large room for improvement, in contrast to popular benchmarks such as N-Cars~\cite{hats} and N-Caltech101~\cite{n_caltech}.
We also experimentally justify the effectiveness of N-ImageNet pretraining.
Models pretrained on N-ImageNet show a large amount of performance gain in various object recognition benchmarks, and are capable of rapidly generalizing to new datasets even with a small number of training samples.
 
We further analyze the robustness of event-based object recognition algorithms amidst changes in camera trajectories and lighting conditions.
Event cameras can operate in highly dynamic scenes and low light environments, but events produced in such conditions tend to have more noise and artifacts from motion blur~\cite{v2e}.
We record variants of N-ImageNet under diverse camera trajectories and lighting, and quantify the significant performance degradation of event-based classifiers under environment changes. 
To the best of our knowledge, our dataset is the first event camera dataset capable of providing quantitative benchmarks for robust event-based object recognition, as shown in Table~\ref{b_as_b}.
In addition, we propose a simple event representation, called DiST (Discounted Sorted Timestamp Image), that shows improved robustness under the external variations.
DiST penalizes events that are more likely to be noise, and uses sorted indices of event timestamps to ensure durability against speed change. 

To summarize, our main contributions are (i) N-ImageNet, the largest fine-grained event camera dataset to date, thus serving as a challenging benchmark, (ii) N-ImageNet pretraining which leads to considerable performance improvement, (iii) N-ImageNet variants that enable quantitative robustness evaluation of event-based object recognition algorithms, and (iv) an event camera representation exhibiting enhanced robustness in diverse environment changes.

\begin{table}[]
\centering
\resizebox{\columnwidth}{!}{
\begin{tabularx}{1.28\columnwidth}{l|cccc}
\toprule
Dataset & \begin{tabular}[c]{@{}c@{}}\# of \\ Samples\end{tabular} & \begin{tabular}[c]{@{}c@{}}\# of \\ Classes\end{tabular} & \begin{tabular}[c]{@{}c@{}}Top \\ Accuracy\end{tabular} & \begin{tabular}[c]{@{}c@{}}Robustness\\ Quantifiable?\end{tabular} \\ \midrule
N-Cars~\cite{hats} & 24029 & 2 & 95.8~\cite{hats} & $\bigtimes$ \\ 
N-Caltech101~\cite{n_caltech} & 8709 & 101 & 90.6~\cite{vid2e} & $\bigtimes$ \\ 
CIFAR10-DVS~\cite{cifar_dvs} & 10000 & 10 & 69.2~\cite{cnn_spiking} & $\bigtimes$ \\ 
ASL-DVS~\cite{asl_dvs} & 100800 & 24 & 94.6~\cite{asl_best} & $\bigtimes$ \\ 
N-MNIST~\cite{n_caltech} & 70000 & 10 & 99.2~\cite{asl_best} & $\bigtimes$ \\ 
MNIST-DVS~\cite{mnist_dvs} & 30000 & 10 & 99.1~\cite{asl_best} & $\bigtimes$ \\ 
N-SOD~\cite{n_sod} & 189 & 4 & 97.14~\cite{n_sod} & $\bigtimes$ \\ 
DVS128-Gesture~\cite{dvs_gesture} & 1342 & 11 & 99.62~\cite{dvs_gesture_best} & $\bigtimes$ \\ 
N-ImageNet & \textbf{1781167} & \textbf{1000} & \textbf{48.93} & \boldmath$\bigcirc$ \\ \bottomrule
\end{tabularx}
}
\caption{Comparison of N-ImageNet with other existing benchmarks for event classification.}
\label{b_as_b}
\end{table}

\section{Related Work}
\paragraph{\textbf{Event Camera Datasets}}
With rising interests in event-based vision, the field has seen a wide range of event camera datasets targeted towards various computer vision tasks, such as object detection~\cite{megapixel, gen1}, optical flow estimation~\cite{evflownet,opt_flow_dataset}, and image reconstruction~\cite{img_reconstruction, img_recon_1, img_recon_2, img_recon_3, img_recon_4, img_recon_5, img_recon_6}. 

For event-based object recognition in particular, diverse datasets~\cite{n_caltech, mnist_dvs, cifar_dvs, asl_dvs, hats, pred18, caltech_256, dvs_gesture, det, roshambo17, sl_animals_dvs, aad_dataset,n_sod} have been proposed, and can be categorized depending on whether the recordings consist of real-world objects or monitor-displayed images.
N-Cars~\cite{hats} and ASL-DVS~\cite{asl_dvs} contain event data obtained by directly capturing various real-world objects.
Such datasets typically have a  smaller number of labels compared to datasets acquired from monitor-displayed images since it is difficult to acquire fine-grained labels of real-world recordings.

Datasets such as N-MNIST, N-Caltech101~\cite{n_caltech}, MNIST-DVS~\cite{mnist_dvs} and CIFAR10-DVS~\cite{cifar_dvs}, which belong to the latter category, are recorded by moving an event camera around monitors displaying images of well-known datasets like MNIST~\cite{mnist} and Caltech101~\cite{caltech101}.
Monitor-generated event camera datasets can be considered synthetic in some aspects, but contain abundant labels from their original image datasets, which are beneficial for training and evaluating event-based object recognition algorithms.
We follow the data acquisition procedure of these datasets, and generate a large-scale, fine-grained dataset by transforming ImageNet~\cite{imagenet} data to event camera recordings.
Furthermore, our experiments demonstrate that leveraging such a large-scale event dataset greatly improves the performance of event-based object recognition algorithms.

\paragraph{\textbf{Event-Based Object Recognition}}
Event camera data exhibits unique characteristics, namely asynchronicity and sparsity.
Existing event-based object recognition algorithms could be classified by whether such characteristics are utilized.
A large body of literature proposes models that perform asynchronous updates in a sparse manner~\cite{asy_sp_1, asy_sp_2, asy_sp_3, asy_sp_4, asy_sp_5, dvs_gesture, hots, hats, asynet, matrix_lstm}.
Recently proposed MatrixLSTMs~\cite{asynet} exhibit competitive results when evaluated on popular event-based object recognition benchmarks such as N-Cars~\cite{hats} and N-Caltech101~\cite{caltech101}.
MatrixLSTMs handle streams of event data using sparse updates of LSTMs~\cite{lstm}, where an adaptive `grouping' operation is used to update the network outputs asynchronously.

Algorithms that avoid the direct exploitation of the sparse and asynchronous nature of event data also prevail~\cite{est, ev_gait, vid2e, gesture_1, graft, unsupervised_feature, asl_dvs}.
These methods place more weight on performance, typically surpassing their sparse and asynchronous competitors in accuracy when tested on various object recognition datasets.
Event Spike Tensors (EST)~\cite{est, vid2e} aggregate events using a learned kernel, resulting in a highly versatile representation of event data.
Relatively simple encodings of event camera data have also been proposed, as in EV-FlowNet~\cite{evflownet} and EV-Gait~\cite{ev_gait}, where events are accumulated to form a four-channel image consisting of event counts and the newest timestamps of each pixel.
We evaluate the performance of the aforementioned object recognition algorithms on N-ImageNet. 
Due to its scale and label diversity, N-ImageNet is capable of providing reliable assessments on different event-based object recognition algorithms.

\paragraph{\textbf{Robustness in Event-Based Object Recognition}}
Event cameras are known to successfully function in low-light conditions and dynamic scenes.
However, the robustness of event-based classifiers under such conditions is a relatively unstudied problem.
Most existing works~\cite{denoise_exp_special, megapixel, hats} either only display qualitative results or experiment with synthetic adversaries.
Sironi \etal~\cite{hats} shows the robustness of their proposed event representation in various real-world objects, but the analysis is only made qualitatively.
Wu \etal~\cite{denoise_exp_special} quantitatively investigates the effect of noise on event-based classifiers, but the experiments are conducted on synthetic noise.
Although there exist previous works such as Wang \etal~\cite{ev_gait} where real event camera noise is investigated, the experiments are carried out with a static camera under constant, ambient lighting.
The N-ImageNet variants recorded under diverse lighting and camera trajectories enable realistic,  quantitative assessment on the robustness of event-based object recognition algorithms.
\section{Method Overview}
\subsection{Dataset Acquisition}
\label{sec:dataset}

\paragraph{\textbf{N-ImageNet Dataset}}
Following the footsteps of previous image-to-event conversion methods~\cite{n_caltech, cifar_dvs, mnist_dvs}, we acquire N-ImageNet from an event camera that observes monitor-displayed images from ImageNet~\cite{imagenet}.
Since events are triggered by pixel intensity changes, external stimuli are solicited to generate event data.
One viable solution for generating constant stimuli would be to keep the camera \emph{still} and make the displayed images \emph{move}, as in~\cite{cifar_dvs, mnist_dvs}.
However, as pointed out by Orchard \etal~\cite{n_caltech}, such methods suffer from artifacts induced by the refresh mechanisms of monitors.
Removing such artifacts requires an additional post-processing step in the frequency domain~\cite{cifar_dvs,mnist_dvs}, which is costly due to the immense number of images in ImageNet, and may alter the inherent subtleties in the raw measurements.

Instead, we opt to \emph{move} the event camera around an LCD monitor displaying \emph{still} images from ImageNet~\cite{imagenet}, as proposed in Orchard \etal~\cite{n_caltech}.
We devise custom hardware to trigger perpetual camera motion as shown in Figure~\ref{device}.
The device consists of two geared motors connected to a pair of perpendicularly adjacent gear racks where the upper and lower motors are responsible for vertical and horizontal motion, respectively.
Each motor is further linked to a programmable Arduino board~\cite{arduino}, which can control the camera movement.
In all our experiments, the event camera is vibrated vertically and horizontally on a plane parallel to the LCD monitor screen.
The amplitude and frequency of vibration are controlled by the program embedded in the Arduino microcontroller~\cite{arduino}.

Once the device is prepared, the event camera is mounted at the forefront of the device for recording event data.
We use the $480\times 640$ resolution Samsung DVS Gen3~\cite{gen3} event camera for recording event sequences, and a 24-inch Dell P2419H LCD monitor for displaying RGB images from ImageNet~\cite{imagenet}.
The acquisition process is performed in a sealed chamber, to ensure that no external light will adulterate the recorded events.
Under this setup, both the training and validation splits of ImageNet are converted to N-ImageNet.
Thanks to the large scale and label granularity of ImageNet, N-ImageNet serves as a challenging benchmark for event-based object recognition and significantly boosts the performance of event-based classifiers via pre-training.
These amenable properties of N-ImageNet will be further examined in Section~\ref{performance}.

\begin{figure}
\centering
\begin{subfigure}{.7\columnwidth}
  \includegraphics[width=.9\columnwidth, left]{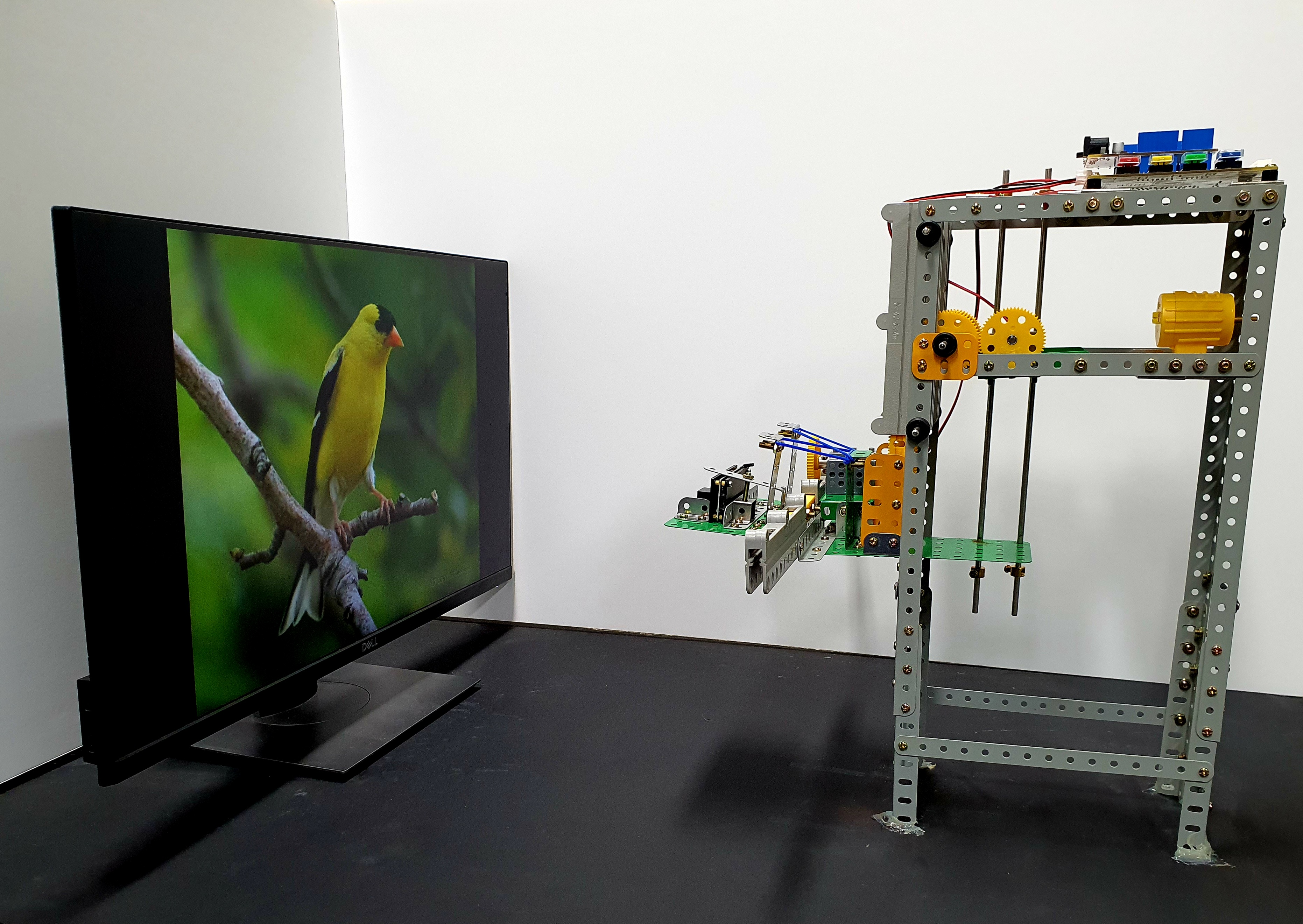}
  \label{fig:sub1}
\end{subfigure}%
\begin{subfigure}{.3\columnwidth}
  \centering
  \includegraphics[height=3.75cm, right]{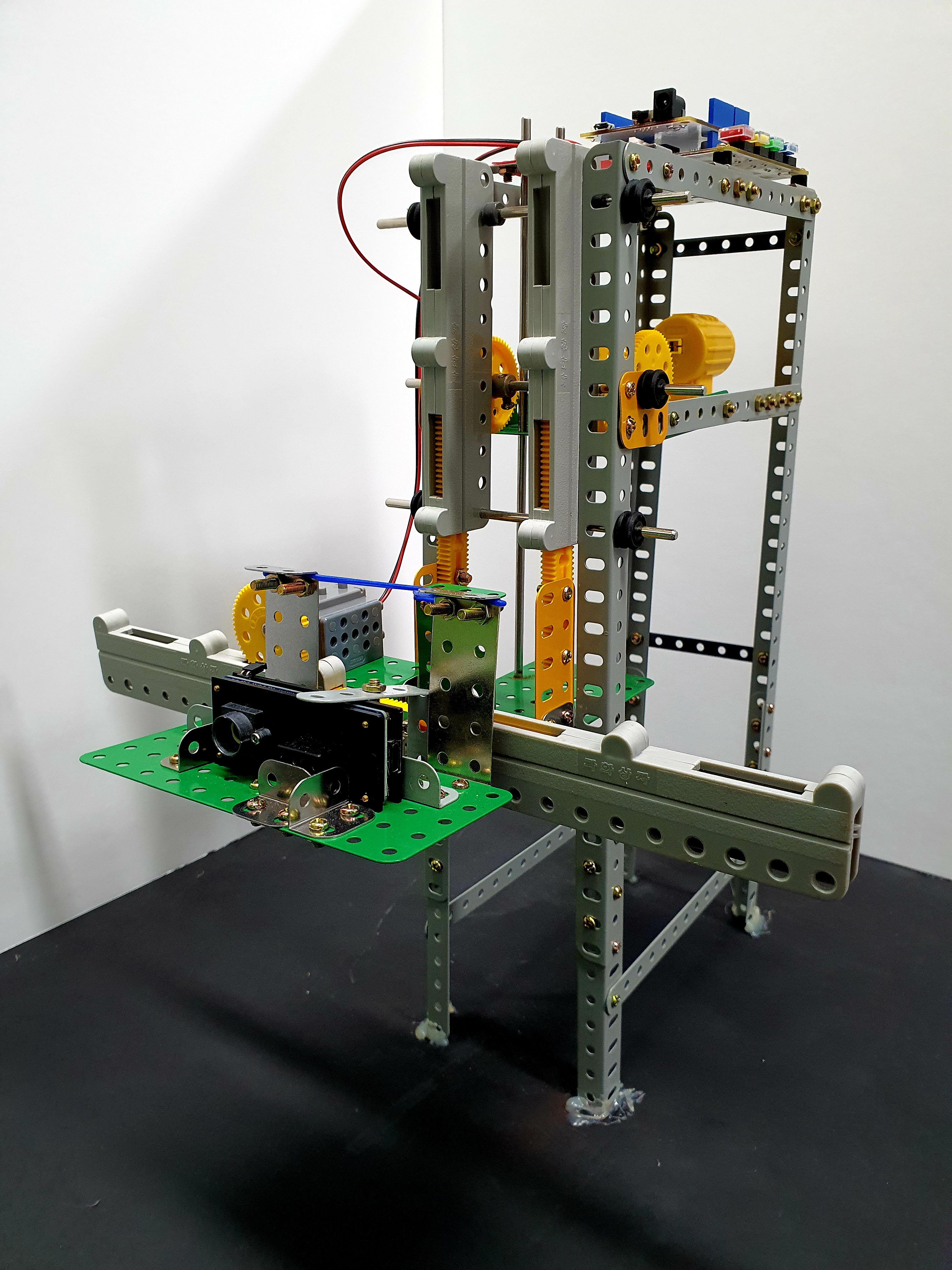}
  \label{fig:sub2}
\end{subfigure}
\vspace{-1.2em}
\caption{Custom hardware designed to convert RGB images to event camera data.}
\label{device}
\end{figure}

\paragraph{\textbf{Datasets for Robustness Evaluation}} 
\label{robust_datasets}
We additionally present a benchmark to quantitatively assess the robustness of event-based object recognition algorithms.
Event representations are vulnerable to alternations in camera motion or illumination, as even small changes can trigger a wide variety of event sequences. 
We simulate such changes using the programmable camera trajectory and monitor brightness from our hardware setup, and generate variants of the N-ImageNet validation split.
We use these validation datasets to quantitatively evaluate the performance degradation of existing object recognition algorithms in Section~\ref{sec:exp_robustness}.

Specifically, we present nine validation datasets to test the robustness of event-based object recognition algorithms amidst changes in motion or illumination.
Table~\ref{tb:traj} lists five datasets with different camera trajectories.
The frequency, amplitude, and trajectory shape of the camera movement are modified with the Arduino microcontroller~\cite{arduino}.
Table~\ref{tb:brightness} shows the monitor configurations of four additional validation datasets, designed to examine the effect of scene brightness changes on event-based classifiers.
Note that Validation 6 and 9 datasets are intended to model scenes with exceedingly low/high illumination by using extreme monitor gamma values.
We also report the illuminance measured at the position of the event camera since the same numerical configurations of different monitors may yield distinct displayed results.
In all cases, other configurations are kept the same as the original N-ImageNet dataset.

\begin{table}[]
\centering
\resizebox{\columnwidth}{!}{
\begin{tabularx}{1.12\columnwidth}{l|ccc}
\toprule
Dataset & Frequency (Hz) & Amplitude (mm) & Shape      \\ \midrule
Original     & 5       & 3       & Square $\circlearrowleft$    \\
Validation 1 & 8.33    & 4.5       & Vertical   \\
Validation 2 & 5       & 3       & Horizontal    \\
Validation 3 & 5       & 6     & Vertical    \\
Validation 4 & 5       & 6       & Horizontal    \\
Validation 5 & 5       & 6       & Square $\circlearrowleft$ \\
\bottomrule

\end{tabularx}
}
\caption{Validation datasets made with various camera motion. $\circlearrowleft$ indicates counterclockwise rotation. The amplitudes of square trajectories represent the lengths of the diagonals.}
\label{tb:traj}
\end{table}

\begin{table}[]
\centering
\resizebox{\columnwidth}{!}{%
\begin{tabularx}{1.1\columnwidth}{l|ccc} \toprule
Dataset & Brightness Level & Gamma & Illuminance (lux) \\ \midrule
Original     & 50                    & 1     & 70.00              \\
Validation 6 & 0                       & 0.7     & 12.75           \\
Validation 7 & 0                   & 1     & 23.38           \\
Validation 8 & 100                   & 1     & 95.50           \\
Validation 9 & 100                   & 1.5     & 111.00           \\ \bottomrule
\end{tabularx}%
}
\caption{Validation datasets generated under various brightness conditions.}
\label{tb:brightness}
\end{table}

\subsection{Robust Event-Based Object Recognition}
\label{robust}
As many event-based classifiers are typically trained and tested in datasets captured in predefined conditions~\cite{n_caltech, asl_dvs}, the performance degradation is inevitable in challenging scenarios that arise in real-life applications.
In Section~\ref{sec:exp_robustness} we evaluate the robustness of existing object recognition algorithms with N-ImageNet variants and demonstrate that external changes indeed incur performance degradation.
Even the best-performing event-based algorithms~\cite{est, matrix_lstm} are fragile to diverse motion and illumination variations.
Ironically, the main benefits of event cameras include the fast temporal response and high dynamic range.

We introduce Discounted Sorted Timestamp Image (DiST), which is designed specifically for robustness against changes in camera trajectory and lighting.
The intuition behind DiST is twofold: (i) noisy events incurred from severe illumination can be suppressed with the evidence of the spatio-temporal neighborhood, and (ii) relative timestamps are robust against the speed of camera motion compared to raw timestamp values.

A typical output of an event camera is a sequence of events $\mathcal{E} = \{e_i=(x_i, y_i, t_i, p_i)\}$, where $e_i$ indicates brightness change with polarity $p_i \in \{-1, 1\}$ at pixel location $(x_i, y_i)$ at time $t_i$.
Given an event camera of spatial resolution $H \times W$, let $\mathcal{N}_{\rho}(x, y, p)$ denote the set of events in $\mathcal{E}$ of polarity $p$, confined within a spatial neighborhood of size $\rho$ around $(x, y)$.
For example, $\mathcal{N}_0(x, y, p)$ would indicate the set of all events of polarity $p$ at the pixel coordinate $(x, y)$.

DiST aggregates a sequence of events $\mathcal{E}$ into a 2 channel image $\mathbf{S} \in \mathbb{R}^{H \times W \times 2}$.
DiST initiates its representation as the timestamp image~\cite{timestamp_image}.
The timestamp image is a 2 channel image, which stores the raw timestamp of the newest event at each pixel, namely $\mathbf{S}_{o}(x, y, p) =  T_\mathrm{new}(\mathcal{N}_{0}(x, y, p))$, where $T_\mathrm{new}(\cdot)$ indicates the newest timestamp.

We first define the Discounted Timestamp Image (DiT) $\mathbf{S}_D$, obtained by subtracting the event occurrence period of the neighborhood from the newest timestamp in $\mathbf{S}_o$,
\begin{equation}
    \mathbf{S}_{D}(x, y, p) = \mathbf{S}_{o}(x, y, p) - \alpha  \mathbf{D}(x, y, p).
\end{equation}
Here $\alpha$ is a constant discount factor and $\mathbf{D}(x,y,p)$ is the neighborhood event occurrence period,
\begin{equation}
\label{disc_amount}
    \mathbf{D}(x, y, p) = \frac{T_\mathrm{new}(\mathcal{N}_{\rho}(x, y, p)) - T_\mathrm{old}(\mathcal{N}_{\rho}(x, y, p))}{C(\mathcal{N}_{\rho}(x, y, p))},
\end{equation}
where $T_\mathrm{old}(\cdot)$ is defined similarly as $T_\mathrm{new}(\cdot)$ and $\rho > 1$.
For each pixel, the discount $\mathbf{D}(x,y,p)$ represents the time range $(T_\mathrm{new}(\cdot)-T_\mathrm{old}(\cdot))$ in which events are generated from the neighborhood $\mathcal{N}_\rho$, normalized by its event count $C(\cdot)$.

\begin{figure}
\centering
\includegraphics[width=\columnwidth]{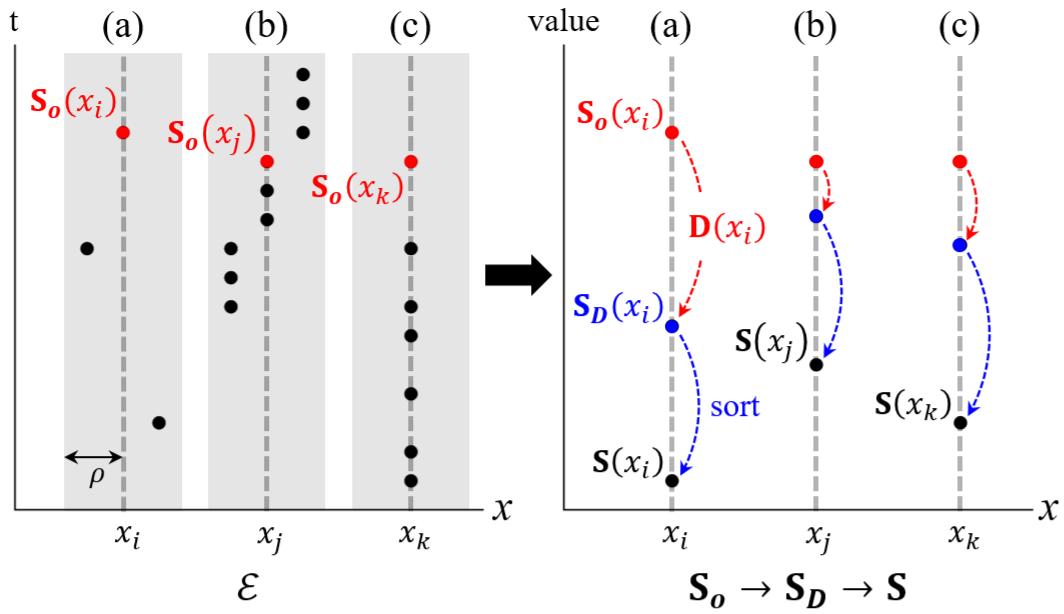}
\caption{Suppressing event camera noise with the discount mechanism. One-dimensional events with a single polarity are shown. Events colored in red are the newest event of each pixel and the neighborhood regions $\mathcal{N}_\rho$ are shaded in gray. (a), (b), and (c) denote background activity, normal events, and hot pixels, respectively. Both the background activity and hot pixels in the raw event sequence (left) result in large discounts and are suppressed by DiST (right).}
\label{fig:noise_invar}
\end{figure}

The discount mechanism is designed to be robust against event camera noise.
Figure~\ref{fig:noise_invar} illustrates the typical patterns for event sequences (left) and the resulting representation of DiST that resolves the discrepancies due to noise (right) in the 1-D case.
Two dominant factors of event camera noise are background activities and hot pixels~\cite{v2e, density_filter}.
Background activities are low frequency noise~\cite{density_filter} more likely to occur in low-light conditions, caused by  transistor leak currents or random photon fluctuations~\cite{v2e, background_activity, o_n_filter}.
Figure~\ref{fig:noise_invar} (a) indicates background activities, whose low frequency results in higher discounts from Equation~\ref{disc_amount}.
Hot pixels are triggered by the improper resets of events~\cite{v2e, density_filter}, and are often spatially isolated~\cite{density_filter} (Figure~\ref{fig:noise_invar} (c)).
Such pixels have a small neighborhood count $C(\cdot)$ and thus are highly discounted in  Equation~\ref{disc_amount}.

The final representation of DiST, $\mathbf{S}$ is the normalized sorted indices of $\mathbf{S}_D$,
\begin{equation}
    \mathbf{S} = \texttt{\small argsort} (\mathbf{S}_{D}) / \max_{x, y, p}~ \texttt{\small argsort} (\mathbf{S}_{D}),
33\end{equation}
where \texttt{\small argsort} denotes the operation that returns the sorted indices of a tensor.
The transformation from $\mathbf{S}_D$ to $\mathbf{S}$ is similar to Alzugaray \etal~\cite{ace}.
However, DiST performs a global sort where all pixels in $\mathbf{S}_D$ are sorted with a single ordering scheme, instead of the local, patch-wise sorting in Alzugary \etal~\cite{ace}.
The global sort allows for an efficient implementation of DiST, which is further explained in the supplementary material.

\begin{figure}[t!]
\centering
    \begin{subfigure}[t]{0.48\linewidth}
    \centering
    \includegraphics[width=\linewidth]{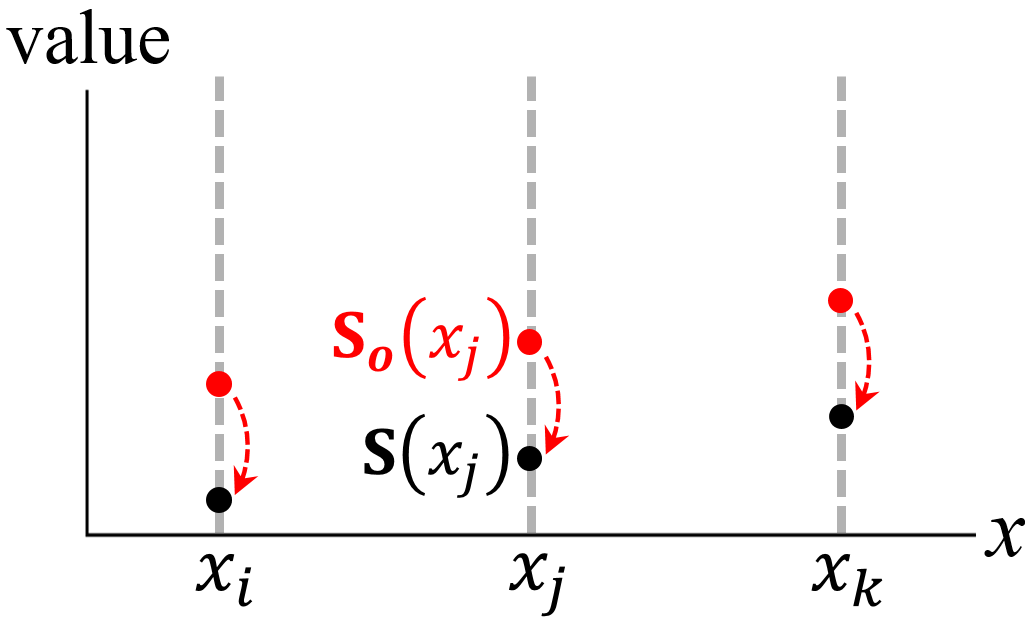}
    \caption{Fast camera motion.}
    \label{fig:fast_motion}
    \end{subfigure}
    ~
    \begin{subfigure}[t]{0.48\linewidth}
    \centering
    \includegraphics[width=\linewidth]{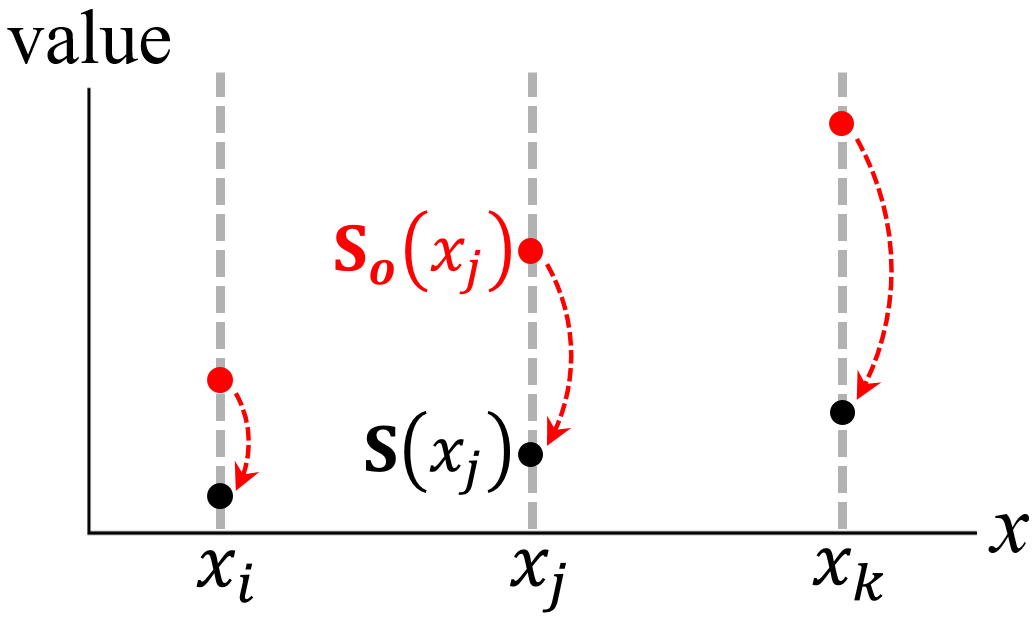}
    \caption{Slow camera motion.}
    \label{fig:slow_motion}
    \end{subfigure}

\caption{Robustness of DiST against event camera speed. Similar to Figure~\ref{fig:noise_invar}, one-dimensional events are displayed and red events denote the newest event. While the timestamps of red events vary with the camera speed, their relative timestamps obtained from sorting remain constant.}
\label{fig:speed_invar}
\vspace{-0.5em}
\end{figure}

By using the normalized relative timestamps, i.e., sorted indices, DiST is resilient against the camera speed.
To illustrate, consider the sequences of one-dimensional events $(x, t, p)$ displayed in Figure~\ref{fig:speed_invar}.
While the absolute values of the timestamps~\cite{timestamp_image} are directly affected by the speed change, DiST remains constant.

We expect DiST to serve as a baseline representation for robust event-based object recognition.
Its robustness against camera trajectory and scene illumination changes will  be quantitatively investigated in Section~\ref{sec:exp_robustness}.

\section{Experimental Results}
\label{exp}

In this section, we empirically validate various properties of N-ImageNet.
With its large scale and label diversity, N-ImageNet is not only a useful benchmark to assess various event-based representations, but can also boost the performance of existing algorithms via pre-training (Section~\ref{performance}).
In Section~\ref{sec:exp_robustness}, we investigate the robustness of event-based classifiers against diverse external conditions, along with the efficacy of our proposed event representation, DiST.

\paragraph{Event Representations for Object Recognition}
We introduce the event representations used throughout our experiments.
The representations are inputs to the object recognition algorithms, while the backbone classifier is fixed to ResNet34~\cite{resnet}.
This is because most event-based object recognition algorithms~\cite{est, matrix_lstm} only differ in the input event representation and share a similar classification backbone.
Eleven event representations are selected for evaluation on N-ImageNet and its variants, as shown in Table~\ref{val_results}.

Two of the representations are learned from the data, namely MatrixLSTM~\cite{matrix_lstm} and Event Spike Tensor (EST)~\cite{est}.
After the events are passed through LSTM~\cite{lstm} for MatrixLSTM and multilayer perceptrons for EST, the outputs are further voxelized to form an image-like representation.

The remaining representations can be classified based on how the timestamps are handled.
Two of these representations discard temporal information, and only use the locations of events.
Binary event image~\cite{binary_image_2, gesture_1} is generated by assigning 1 to pixels with events, and 0 to others.
Event histogram~\cite{event_driving} is an extension of the former, additionally keeping the event count of each pixel.

Three representations use the raw timestamps to leverage temporal information.
Timestamp image~\cite{timestamp_image} caches the newest timestamp for each pixel.
Event image~\cite{ev_gait, evflownet} is a richer representation that concatenates the event histogram~\cite{event_driving} and timestamp image~\cite{timestamp_image}.
Time surface~\cite{hots} extends the timestamp image~\cite{timestamp_image} in a slightly different manner, by passing each timestamp through an exponential filter.
This allows the surface to place more weight on the newest events, which enhances the sharpness of the representation.
The aforementioned representations with raw timestamps can be vulnerable to camera speed changes, as pointed out in Section~\ref{robust}.

We further include representations targeted to enhance robustness.
HATS~\cite{hats} improves the robustness against event camera noise.
Specifically, the outliers are smoothed by aggregating neighboring pixels of the time surface~\cite{hots}.
We use a slightly modified version of HATS~\cite{hats} for more competitive results, and the details are provided in the supplementary material.
Surface of active events with sort normalization~\cite{ace}, which we will refer to as sorted time surface, is robust against camera speed changes as the sorting generates relative timestamps. 
DiST, as explained in Section~\ref{robust}, is robust against both event camera noise and speed changes.
We additionally report results on the variant of DiST without sorting, namely the Discounted Timestamp Image (DiT).
Evaluation on DiT can shed light on the importance of the sorting operation in DiST.

\paragraph{Implementation Details}
All inputs are reshaped into a $224 \times 224$ grid to restrict GPU memory consumption and shorten inference time.
All models are trained from scratch with a learning rate of $0.0003$, except for the learned representations (MatrixLSTM~\cite{matrix_lstm} and EST~\cite{est}).
The weights are initialized with ImageNet pre-training for these representations to fully replicate the training setup specified in the original works.
We train these models with a learning rate of $0.0001$.
Further information regarding experimental details is provided in the supplementary material.

\begin{table}[]
\resizebox{\columnwidth}{!}{
\begin{tabular}{l|c|c|c}
\toprule
Representation & Description & \begin{tabular}[c]{@{}c@{}}\# of\\ Channels\end{tabular} & Accuracy(\%) \\ \Xhline{2\arrayrulewidth}
MatrixLSTM~\cite{matrix_lstm} & \begin{tabular}[c]{@{}c@{}}Learned with\\ LSTM\end{tabular} & 3 & 32.21 \\ \hline
\begin{tabular}[c]{@{}l@{}}Event Spike \\ Tensor~\cite{est}\end{tabular} & \begin{tabular}[c]{@{}c@{}}Learned with\\ MLP\end{tabular} & 18 & \textbf{48.93} \\ \hline
\begin{tabular}[c]{@{}l@{}}Binary Event \\ Image~\cite{binary_image_2} \end{tabular} & \begin{tabular}[c]{@{}c@{}}Binarized\\ event occurence\end{tabular} & 2 & 46.36 \\ \hline
\begin{tabular}[c]{@{}l@{}} Event\\ Histogram~\cite{event_driving}
\end{tabular} & Event counts & 2 & 47.73 \\ \hline
Event Image~\cite{ev_gait} & \begin{tabular}[c]{@{}c@{}}Event counts and\\ newest timestamps\end{tabular} & 4 & 45.77 \\ \hline
Time Surface~\cite{hots} & \begin{tabular}[c]{@{}c@{}}Exponential of\\ newest timestamps\end{tabular} & 2 & 44.32 \\ \hline
HATS~\cite{hats} & \begin{tabular}[c]{@{}c@{}}Aggregated\\ newest timestamps\end{tabular} & 2 & 47.14 \\ \hline
\begin{tabular}[c]{@{}l@{}} Timestamp\\ Image~\cite{timestamp_image}
\end{tabular} & \begin{tabular}[c]{@{}c@{}}Newest \\ timestamps\end{tabular} & 2 & 45.86 \\ \hline
\begin{tabular}[c]{@{}l@{}}Sorted Time \\ Surface~\cite{ace}\end{tabular} & \begin{tabular}[c]{@{}c@{}}Sorted newest\\ timestamps\end{tabular} & 2 & 47.90 \\ \hline
DiT & \begin{tabular}[c]{@{}c@{}}Discounted\\ newest timestamps\end{tabular} & 2 & 46.1 \\ \hline
DiST & \begin{tabular}[c]{@{}c@{}}Sorted discounted\\ timestamps \end{tabular} & 2 & \textbf{48.43} \\ \bottomrule
\end{tabular}
}
\caption{N-ImageNet validation accuracy evaluated on various event representations.}
\label{val_results}
\end{table}

\subsection{Evaluation Results with N-ImageNet}

\paragraph{Event-based Object Recognition}
\label{performance}
Table~\ref{val_results} displays the evaluation results of existing event-based object recognition algorithms on N-ImageNet.
The accuracy of the best performing model on N-ImageNet (48.9\%), is far below that of the state-of-the-art model on ImageNet~\cite{imagenet_sota} (90.2\%).
The clear gap indicates that mastering N-ImageNet is still a long way to go.

Other examined models also exhibit a stark contrast in their reported accuracy on existing benchmarks and performance on N-ImageNet.
For example, the test accuracy of the event histogram~\cite{asynet} on N-Cars is 94.5\%, and the test accuracy of MatrixLSTM~\cite{matrix_lstm} on N-Caltech101 is 86.6\%.
These models show a validation accuracy of around $30\sim50\%$ in N-ImageNet, further supporting the difficulty of N-ImageNet.
N-ImageNet is a large-scale, fine-grained benchmark (Table~\ref{b_as_b}) compared to any other existing benchmark and the inherent challenge will foster development in event classifiers that could readily function in the real world.

\paragraph{Assessment on Representations}
The evaluation of various representations on N-ImageNet allows us to make a systematic assessment of different design choices to handle event-based data.
Interestingly, the performance of representations without temporal information (binary event image~\cite{gesture_1,binary_image_2} and event histogram~\cite{event_driving}) are superior to representations directly using raw timestamps (timestamp image~\cite{timestamp_image}, time surface~\cite{hots}, and event image~\cite{ev_gait,evflownet}).
The wide variations in raw timestamps deteriorate the generalization capacity of representations that directly utilize this information.
This notion is further supported by the fact that the representations using relative timestamps (sorted time surface~\cite{ace} and DiST) outperform those using raw timestamps.

It should also be noted that our proposed robust representation, DiST, successfully generalizes to large-scale datasets such as N-ImageNet, and shows performance on par with strong learned representations.
EST~\cite{est} is the best performing model in Table~\ref{val_results}, capable of learning highly expressive encodings of event data, thanks to its event aggregation using multilayer perceptrons.
The performance of DiST is very close to that of EST, although it does not incorporate any learnable module in its event representation.
The suppression of noise from discounting, and the resilience to variations in camera speed from using relative timestamps help DiST to generalize.
If we either omit the discount (sorted time surface~\cite{ace}) or the sorting mechanism (DiT), the performance is inferior to DiST, indicating the importance of the discounting and sorting operations.
We further investigate the robustness of DiST in Section~\ref{sec:exp_robustness}.

\begin{table}[]
\centering
\resizebox{\columnwidth}{!}{%
\begin{tabularx}{1.25\columnwidth}{l|c|c|c|c}
\toprule
Dataset & N-Cars & CIFAR10-DVS & ASL-DVS & N-Caltech101 \\ \midrule
\# of classes & 2 & 10 & 24 & 101 \\ \midrule
Random & 90.80 & 62.57 & 29.57 & 68.12 \\ 
ImageNet & 91.48 & 70.36 & 53.43 & 80.88 \\ 
N-ImageNet & \textbf{94.73} & \textbf{73.72} & \textbf{58.28} & \textbf{86.81} \\ \bottomrule
\end{tabularx}
}
\caption{Test accuracy of N-ImageNet pretrained models on existing event-based object recognition benchmarks, compared with ImageNet pretraining and random initialization.}
\label{major_stats}
\end{table}

\begin{figure}
    \makebox[\columnwidth][c]{
        \includegraphics[width=0.5\columnwidth]{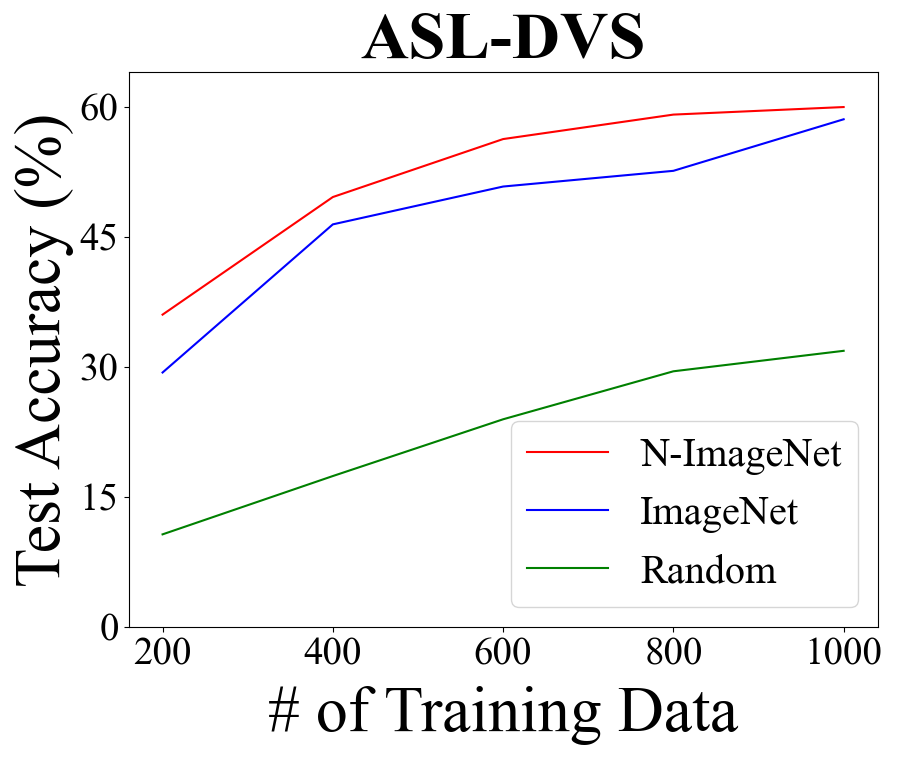}
        \includegraphics[width=0.5\columnwidth]{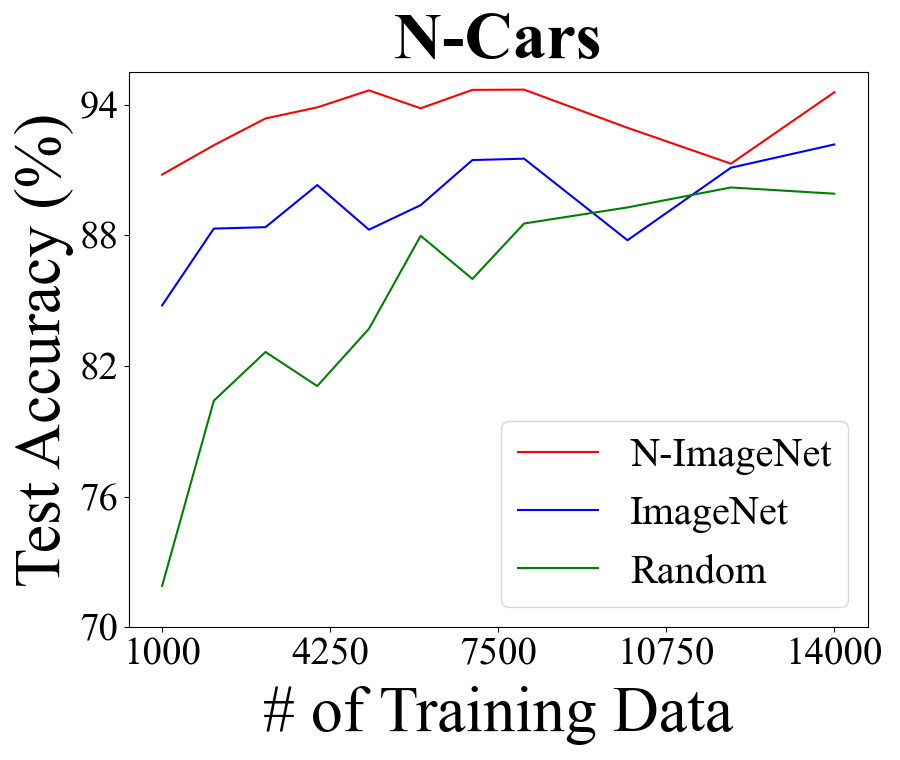}
    }
    \makebox[\columnwidth][c]{
        \includegraphics[width=0.5\columnwidth]{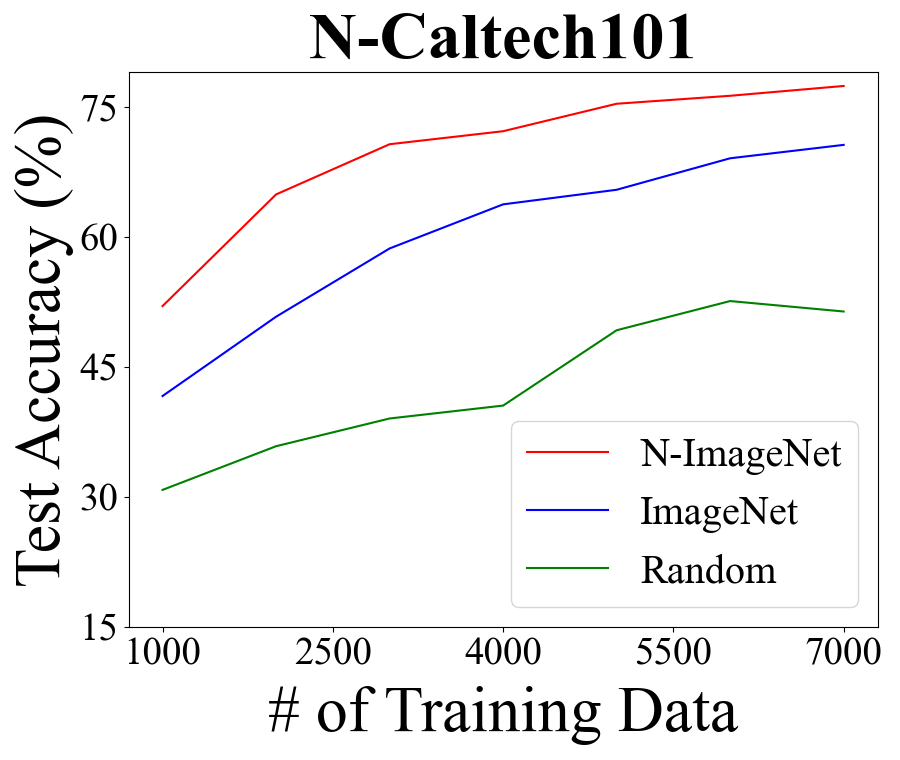}
        \includegraphics[width=0.5\columnwidth]{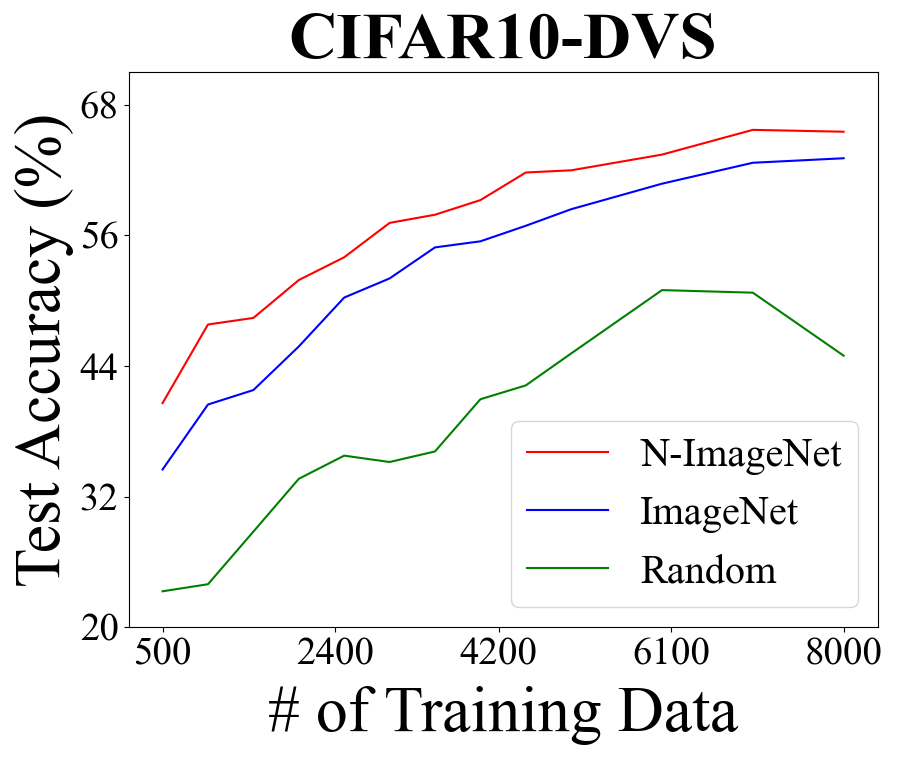}
    }
    \caption{Test accuracy of N-ImageNet pretrained models in resource constrained settings. Each model is trained for 5 epochs with varying amounts of training data.}
    \label{fig:pretrain_four}
\end{figure}

\paragraph{Efficacy of N-ImageNet Pre-Training}
\label{pre-training}
Apart from being a challenging benchmark, the main motivation of N-ImageNet is to provide a large reservoir of event data to pre-train powerful representations for downstream tasks, echoing the role of ImageNet in conventional images.
We validate the effectiveness of N-ImageNet pre-training by observing the capacity to generalize in new, unseen datasets.
Four standard event camera datasets are used for evaluation: N-Caltech101~\cite{n_caltech}, N-Cars~\cite{hats}, CIFAR10-DVS~\cite{cifar_dvs}, and ASL-DVS~\cite{asl_dvs}.
For seven event representations from Table~\ref{val_results}, ResNet34~\cite{resnet} is pre-trained on N-ImageNet and compared with ImageNet pre-training and random initialization.
The seven representations selected are as follows: binary event image~\cite{binary_image_2}, event histogram~\cite{event_driving}, timestamp image~\cite{timestamp_image}, event image~\cite{ev_gait}, time surface~\cite{hots}, sorted time surface~\cite{ace}, and DiST.
In experiments explicated below, we report the averaged test accuracy of the seven representations on each dataset.
Additional details about the experimental setup are specified in the supplementary material.

Table~\ref{major_stats} displays the average test accuracy after training a fixed number of epochs for different initialization schemes.
Note that we only use 800 samples from ASL-DVS~\cite{asl_dvs} for training, as using the whole dataset made all model performances saturate near 99\%.
Networks initiated with N-ImageNet pre-trained weights outperform models from other initialization schemes by a large margin.
Notably, the benefits of pre-training intensify as the number of classes in the datasets increases.
This could be attributed to the fine-grained labels of N-ImageNet, which help models to generalize in challenging datasets where numerous labels are present.
Furthermore, N-ImageNet pre-trained models outperform its competitors in N-Cars and ASL-DVS, which are recordings of real-world objects.
This indicates that although N-ImageNet contains events from monitor displayed images, models pre-trained on it could seamlessly generalize to recognizing real-world objects.

As a practical extension to the previous experiment, we validate the generalization capability of N-ImageNet pre-trained models under resource-constrained settings.
We train the same set of models for 5 epochs with initialization schemes from the previous experiment, under varying numbers of training samples.
Figure~\ref{fig:pretrain_four} shows that N-ImageNet pre-training incurs a large performance improvement across all four evaluated datasets.
The performance gain is further increased when the number of training samples is small.
Such results imply that N-ImageNet pre-training provides strong semantic priors that enable object recognition algorithms to quickly adapt to new datasets, even with a few labeled samples.

\begin{table}[]
\centering
\resizebox{\columnwidth}{!}{
\begin{tabularx}{1.05\columnwidth}{l|c|c|c|c}
\toprule
Factor & \multicolumn{2}{c|}{Trajectory} & \multicolumn{2}{c}{Brightness} \\ \midrule
Change Amount & Small & Big & \multicolumn{1}{c|}{Small} & Big \\ \midrule
\begin{tabular}[c]{@{}l@{}}Validation Dataset\\ Number\end{tabular} & 1, 2 & 3, 4, 5 & \multicolumn{1}{c|}{7, 8} & 6, 9 \\ \midrule
MatrixLSTM~\cite{matrix_lstm} & 33.00 & 25.62 & \multicolumn{1}{c|}{28.91} & 23.60 \\ 
Event Spike Tensor~\cite{est} & 36.97 & 32.35 & \multicolumn{1}{c|}{24.89} & 22.36 \\ 
Binary Event Image~\cite{binary_image_2} & 36.68 & 31.82 & \multicolumn{1}{c|}{30.94} & 25.54 \\ 
Event Histogram~\cite{event_driving} & 38.72 & 32.49 & \multicolumn{1}{c|}{33.01} & 27.72 \\ 
Event Image~\cite{ev_gait} & 36.52 & 30.96 & \multicolumn{1}{c|}{32.26} & 27.04 \\ 
Time Surface~\cite{hots} & 37.82 & 33.46 & \multicolumn{1}{c|}{34.19} & 28.74 \\ 
HATS~\cite{hats} & 38.95 & 33.28 & \multicolumn{1}{c|}{33.26} & 28.22 \\ 
Timestamp Image~\cite{timestamp_image} & 38.31 & 33.70 & \multicolumn{1}{c|}{33.27} & 28.04 \\ 
Sorted Time Surface~\cite{ace} & 38.92 & 33.69 & \multicolumn{1}{c|}{33.47} & 28.38 \\ 
DiT & 38.21 & 33.61 & \multicolumn{1}{c|}{32.66} & 28.42 \\ 
DiST & \textbf{40.88} & \textbf{35.85} & \multicolumn{1}{c|}{\textbf{35.87}} & \textbf{30.88} \\ \bottomrule
\end{tabularx}
}
\caption{Mean accuracy measured on N-ImageNet variants with changes in camera trajectory and brightness.}
\label{degrad_stats}
\end{table}

\subsection{Robust Event-Based Object Recognition}
\label{sec:exp_robustness}

\paragraph{\textbf{Validation Accuracy of N-ImageNet Variants}} 
Using the N-ImageNet variants created under various external conditions as described in Section~\ref{robust_datasets}, we examine the robustness of event-based object recognition algorithms.
Table~\ref{degrad_stats} shows the validation accuracy averaged over the trajectory-modified datasets and brightness-modified datasets.
All models displayed in Table~\ref{val_results} are evaluated on the N-ImageNet variants.
We group datasets according to the variation factor, i.e., brightness and trajectory, and the amount of discrepancy between the original setup and the modified setup.

All tested models exhibit a consistent deterioration in performance when evaluated on the N-ImageNet variants.
Furthermore, the amount of performance degradation intensifies as the amount of environment change increases, as shown in Table~\ref{degrad_stats}.
These observations imply that many event-based object recognition algorithms are biased on their training setups, and thus fail to fully generalize in challenging, unseen environments.
In spite of the consistent performance drop however, DiST outperforms its competitors under all external variations shown in Table~\ref{degrad_stats}.
Notably, the ablated versions of DiST, i.e. sorted time surface~\cite{ace}, DiT, and timestamp image~\cite{timestamp_image}, all perform poorly compared to DiST.
Along with the validation accuracy on the original N-ImageNet, this reinforces the necessity of both the discounting and sorting modules of DiST.
DiST's capacity to generalize in unseen environmental conditions demonstrates its effectiveness as a robust representation for event-based object recognition.

\paragraph{\textbf{Representation Consistency}} To further investigate the robustness of DiST, we quantify the content-wise consistency of various event representations.
Seven representations from Table~\ref{val_results} are compared against DiST.
Other three representations (MatrixLSTM~\cite{matrix_lstm}, EST~\cite{est}, event image~\cite{ev_gait}) are omitted as they have different number of channels, which may incur unfair comparison.
For each representation, we assess the structural similarity index measure (SSIM) between the original representation from N-ImageNet and the representation obtained from N-ImageNet variants.
To further elaborate, suppose $\mathcal{E}_\mathrm{orig}$ and $\mathcal{E}_\mathrm{var}$ are event sequences derived from the same image in the ImageNet validation dataset.
Let $\mathbf{R}_\mathrm{orig}$ and $\mathbf{R}_\mathrm{var}$ be the event representations obtained from $\mathcal{E}_\mathrm{orig}$ and $\mathcal{E}_\mathrm{var}$ respectively.
We report $\text{SSIM}(\mathbf{R}_\mathrm{orig}, \mathbf{R}_\mathrm{var})$, which measures how consistent each representations are amidst external condition changes.

\begin{figure}
\centering
\includegraphics[width=0.9\columnwidth]{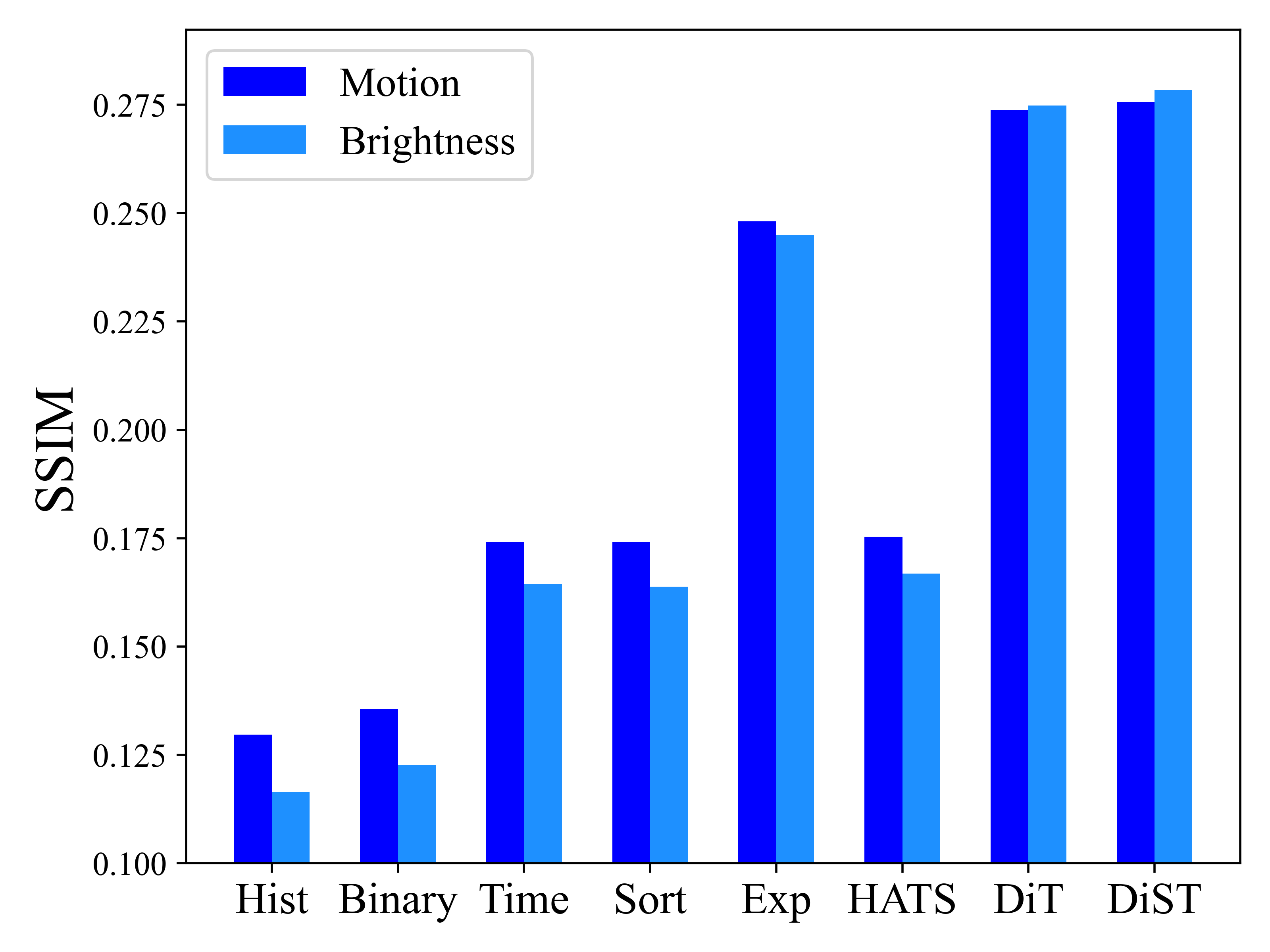}
\caption{Structural similarity measure (SSIM) between the representations from N-ImageNet and its variants, grouped by changes in motion and brightness. 
High SSIM indicates that the structure of the representation is stable under external variations.
Note that `Time' and `Exp' denote timestamp image~\cite{timestamp_image} and time surface~\cite{hots}, respectively.
}
\label{fig:ssim}
\end{figure}

As displayed in Figure~\ref{fig:ssim}, the contents of DiST are more consistent than other competing representations, which can be observed from its highest SSIM value.
The contribution of discounting is greater than that of sorting in representation consistency, which can be seen from the SSIM difference of DiT and DiST.
However, sorting is crucial for robust object recognition, as can be observed from Table~\ref{degrad_stats}, where a clear gap exists between DiT and DiST.
Thus, the interplay between discounting and sorting as a whole enhances the robustness of DiST, further leading to improved performance in N-ImageNet variants.

Apart from the robustness of DiST, it must be noted that the N-ImageNet variants serve as the first benchmark for quantifying robustness in event-based classifiers.
Although DiST shows a consistent improvement from previous models in robustness, it does not fully recover the original N-ImageNet validation accuracy reported in Table~\ref{val_results}.
We expect the N-ImageNet variants to spur future work in robust representations for event-based object recognition.

\section{Conclusion}
In this paper, we introduce N-ImageNet, a large-scale dataset for robust, fine-grained object recognition with event cameras.
The performance of numerous event-based classifiers on N-ImageNet demonstrates the potential of our dataset as a challenging benchmark.
N-ImageNet pretraining is empirically proven to be beneficial, boosting the performance of many existing object recognition algorithms, even under resource-constrained settings. 
Assessments carried out on the variants of N-ImageNet enable the quantitative evaluation of event classifiers' robustness, and demonstrate the bias present in many event-based object recognition algorithms.
As a first step towards remedying such biases, a novel representation, DiST, is proposed that outperforms all the tested models in N-ImageNet variants.
We expect N-ImageNet to foster the development of event-based object recognition algorithms that could readily function in real-life applications.

\paragraph{Acknowledgments} This work was supported by by the National Research Foundation of Korea (NRF) grant funded by the Korea government (MSIT) (No. 2020R1C1C1008195), Samsung Electronics Co., Ltd, and the BK21 FOUR program of the Education and Research Program for Future ICT Pioneers, Seoul National University in 2021.

\bibliographystyle{ieee}
\bibliography{n_imagenet_bib}

\end{document}